\documentclass[11pt,a4paper]{article}
\usepackage[hyperref]{emnlp2018}
\usepackage{times}
\usepackage{latexsym}

\usepackage{url}

\aclfinalcopy 

\usepackage{times,latexsym,arydshln,amsmath,graphicx,xcolor,multirow,arydshln}
\usepackage{amsfonts, amssymb}

\newcommand{\todooff}{\long\gdef\todo##1{}}
\newcommand{\todoon}{\long\gdef\todo##1{{
\bf\textcolor{red} {TODO: ##1}
}}}
\todoon
\todooff

\newcounter{notecounter}
\newcommand{\enotesoff}{\long\gdef\enote##1##2{}}
\newcommand{\enoteson}{\long\gdef\enote##1##2{{
\stepcounter{notecounter}
{\large\bf
\hspace{1cm}\arabic{notecounter} $<<<$ ##1: ##2
$>>>$\hspace{1cm}}}}}
\enoteson
\enotesoff

\def\figref#1{Figure~\ref{fig:#1}}
\def\figlabel#1{\label{fig:#1}\label{p:#1}}

\def\tabref#1{Table~\ref{tab:#1}}
\def\tablabel#1{\label{tab:#1}\label{p:#1}}

\def\secref#1{\S\ref{sec:#1}}
\def\seclabel#1{\label{sec:#1}}
\def\eqref#1{Eq.~\ref{eqn:#1}}

\def\eqlabel#1{\label{eqn:#1}}

\title{Multi-Multi-View Learning:\\ Multilingual and
  Multi-Representation Entity Typing}

\author{Yadollah Yaghoobzadeh \\
  Microsoft Research \\ Montreal, Canada \\
  {\tt yayaghoo@microsoft.com} \\\And
  Hinrich Sch\"utze \\
  LMU Munich \\ Munich, Germany \\
  {\tt inquiries@cislmu.org} \\}

\date{}

\def\mathlinebreak{\\[0.1cm]}
\def\mathindent{\mbox{\hspace{0.5cm}}}

\begin{document}
\maketitle

\begin{abstract}
Knowledge bases (KBs) are paramount
in NLP. We employ
\emph{multiview learning}
for increasing accuracy and
coverage of entity type information in KBs. We rely on two
\emph{metaviews}:
language and representation. For 
language, we consider high-resource and low-resource
languages from Wikipedia. For representation, we
consider representations based 
on the context distribution of the entity (i.e., on its embedding),
on the entity's name (i.e., on its surface form) and on its
description in Wikipedia.
The two metaviews
language and representation can be freely combined:
each pair of language and
representation (e.g., German  embedding, English description,
Spanish name) is a distinct view.
Our experiments on entity typing with fine-grained classes demonstrate the effectiveness of multiview learning.
We release MVET, a large multiview -- and, in particular,
multilingual -- entity typing dataset we created.
Mono- and multilingual fine-grained entity typing systems can be evaluated on this dataset.
\end{abstract}

\section{Introduction}
\seclabel{intro}
Accurate and complete knowledge bases (KBs) are paramount
in NLP.
Entity typing, and in particular fine-grained entity typing, is an important component of KB completion
with applications in NLP and knowledge engineering.  Studies
so far have been mostly for English \cite{figment15}, but
also for Japanese \cite{Suzuki2016et}.

We employ
\emph{multiview learning}
for increasing accuracy and
coverage of entity type information in KBs. We rely on two
\emph{metaviews}:
language and representation. For 
language, we take high- and low-resource
languages from Wikipedia. For representation, we
consider representations based 
on the context distribution of the entity (i.e., on its embedding),
on the entity's name (i.e., on its surface form) and on its
description in Wikipedia.
The two metaviews
language and representation can be freely combined:
each pair of language and
representation (e.g., German embedding, English description,
Spanish name) is a distinct view.

Views 
are defined as kinds of information
about an instance that have three properties \cite{blum98cotraining,xu2013survey}.
(i)
Sufficiency. Each view
is sufficient for
classification on its own. (ii) Compatibility. The target
functions in all
views predict the same labels for cooccurring features
with high
probability. (iii) Conditional independence. The views
are conditionally
independent given the class label.

As in most cases of multiview learning,
these three properties are only approximately true for our problem.
(i) Not every view is sufficient
for every instance. While a name like ``George Washington
Bridge'' is sufficient for typing the entity as ``bridge'',
the name ``Washington'' is not sufficient for entity typing.
(ii) Cases of incompatibility exist. For example, the
``Bering Land Bridge'' is not a bridge. (iii) Views have
some degree of conditional dependence. For example, if a
bridge is a viaduct, not a bridge proper, then the
description of the bridge will contain more occurrences of
the word ``viaduct'' than for proper bridges whose name
does not contain the word ``viaduct''.

In summary, we make three main contributions. (i) 
We formalize entity typing as a multiview
problem by introducing two metaviews, language and representation; each
combination of instances of these two metaviews defines a distinct view.
(ii) We show that this formalization is effective for entity
typing as a key task in KB completion: multiview and crossview learning outperform
singleview learning by a large margin, especially for rare entities 
and low-resource languages.
(iii)
We release MVET (Multiview Entity Typing),  a large multiview and, in particular,
multilingual dataset, for entity typing\footnote{Our dataset and code are available at: \url{http://github.com/yyaghoobzadeh/MVET}}. 
This dataset can be used for mono- and multilingual fine-grained 
entity typing evaluations. 
In contrast to prior work on entity typing based on Clueweb (a commercial corpus), all our data can be released publicly because it is based on Wikipedia.

\begin{figure}
  \centering
\includegraphics[scale=0.38]{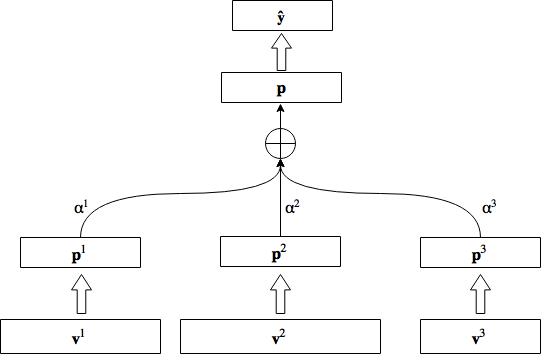}
\caption{Attention-based multiview learning.
View specific representations  $\mathbf{v}^j$ of the entity are  transformed to 
a shared space and summed by attention weights $\alpha^j$
into aggregated multiview representation $\mathbf{p}$.
A one-hidden-layer perceptron computes  output
vector $\mathbf{\hat{y}}$.}
\figlabel{att}
\end{figure}

\section{Multilingual Multi-Representation Entity Typing}
\seclabel{multilingual} We address the task of
entity typing \cite{figment15}, i.e., assigning to a given entity one or more types
from a set of types $T$. E.g.,
Churchill is a politician and writer.

Our key idea is that we can tap two different
information sources for entity typing, which we
will refer to as \emph{metaviews}: language and
representation.  For the language metaview, we
consider $N$ languages (English, German, \ldots).
For the representation metaview, we consider three
representations: based on the entity's context
distribution, based on its canonical name and
based on its description.  Each combination of
language and representation defines a separate
view of the entity, i.e., we have up to $3N$
views.  The views are not completely
independent of each other: what is written about
an entity in English and German is correlated and
information derived from the entity name is
correlated with its description; see discussion in
\secref{intro}.  Still, each view contains
information complementary to each other view.

For the \emph{context view} of the representation metaview, we use entity embeddings \cite{figment15}: each mention of
an entity in Wikipedia
-- identified using Wikipedia hyperlinks --
is replaced
by the entity's unique identifier. 
We can then run standard embedding learning. 
For the \emph{name view}, we take the
sum of the embeddings of the words of the entity name.
The \emph{description view} is based on the entity's
Wikipedia page; see \secref{entityrep}.

We represent view $j$ of an entity $e$ as the
vector or embedding $\mathbf{v}^j  \in \mathbb{R} ^{d_j}$.
We combine
these embeddings into a multiview representation
$\mathbf{p} \in \mathbb{R} ^{d}$ of entity $e$.  
As discussed above, each
$\mathbf{v}^j$ contributes potentially
complementary information.

After learning $\mathbf{p}$,
a one-hidden-layer perceptron
computes the
type predictions $\mathbf{\hat{y}}  \in \mathbb{R} ^{|T|}$:
\begin{equation}
\mathbf{\hat{y}} = \sigma\Big(\mathbf{W}_o f\big(\mathbf{W}_h\mathbf{p}\big) \Big)
\eqlabel{mlp}
\end{equation}
$f$ is leaky rectifier,
$\mathbf{W}_h  \in \mathbb{R} ^{h \times d}$,
$\mathbf{W}_o \in \mathbb{R} ^{|T| \times h}$.

The cost function is  binary cross entropy summed over types and training examples:
\begin{equation}
 \sum_{i,t} (\mathbf{y}_{i,t} \log(\mathbf{\hat{y}}_{i,t}) + (1 \!\!-\!\!\mathbf{y}_{i,t}) (\log(1\!\!-\!\!\mathbf{\hat{y}}_{i,t})) )
 \end{equation}
where $\mathbf{y}_{i,t}$ and $\mathbf{\hat{y}}_{i,t}$ are
the gold and prediction for type $t$ of 
example $i$.

A simple and effective way of computing the representation
$\mathbf{p}$ of an entity
is  what we refer to as \textbf{MULTIVIEW-CON}: a
concatenation of the $n$ view embeddings, followed by a
non-linear transformation:
\begin{equation*}
 \mathbf{p} = \tanh(\mathbf{W}_1[\textbf{v}^{1};
  \textbf{v}^{2}; ... ; \mathbf{v}^{n}]^\intercal) 
  \end{equation*}
  where
$\mathbf{W}_1 \in \mathbb{R}^{d \times (\Sigma_{j=1}^{n} d_j) }$ is the transformation matrix.

Concatenation may not be effective because
some entities have pages in all Wikipedias and
thus have  $200 \times 3 =600$ views
whereas others occur only in one.
Also, the views might have 
 different qualities. Therefore, we consider
attention-based weighted average or
\textbf{MULTIVIEW-ATT} as an alternative to MULTIVIEW-CON.
Embeddings $\mathbf{v}^j$ live
in different spaces, so we first transform them
using language specific matrices
$\mathbf{W}^j  \in \mathbb{R} ^{d \times d_j}$:
\begin{equation}\eqlabel{sharedspaceattention}
\mathbf{p}^j = \tanh (\mathbf{W}^j \mathbf{v}^j)
\end{equation}
Then, we compute the attention weights:\mathlinebreak
\mathindent $\alpha^j = \mbox{softmax}(\mathbf{a}^{T} \mathbf{p}^j)$ \mathlinebreak
where $\mathbf{a} \in \mathbb{R}^{d}$ is a vector that is trained  to weight the 
vectors $\mathbf{p}^j$. 
The \textbf{MULTIVIEW-ATT} representation is then defined as:
\begin{equation*}
 \mathbf{p} = \sum_j  \alpha^j \mathbf{p}^j
 \end{equation*}
A schematic architecture is shown in \figref{att}.

We also experiment with two alternatives.
\textbf{MULTIVIEW-AVG}:
We set all $\alpha^j = 1/n$, i.e., the entity representation
is a simple average.
\textbf{MULTIVIEW-MAX}:
We  apply  per-dimension maxpooling,
$\mathbf{p}_i = \max_j   \mathbf{p}^j_i$.
The idea here is to capture the most significant features
across views.

\section{Dataset and Experiments}
In this section, we first introduce our new dataset and then describe our results.

\textbf{Multiview entity typing (MVET) dataset}.
Wikipedia and Freebase are our
sources for creation of MVET.  We try to
map each English Wikipedia article of an entity to Freebase.
Freebase types are mapped to 113 FIGER tags \cite{ling2012fine}. We
use  Wikipedia
interlingual links
to build
multilingual datasets by identifying corresponding Wikipedia
articles in non-English languages. 
So for each entity, we have the English
article name as well as the names in other languages (if they
exist) and FIGER types of the entity.
We use these multilingual names and Wikipedias to build our representation views 
as described in \secref{entityrep}.

We experiment with ten languages: English (EN), German (DE),
Farsi (FA), Spanish (ES); and Arabic (AR), French (FR),
Italian (IT).  Polish (PL), Portuguese (PT), Russian (RU).
The procedure described above gives us around 2M
entities.  We divide them into train (50\%), dev (20\%) and
test (30\%) and, for efficient training, sample them
stratified by type to ensure enough entities per type.  The
final dataset used in our experiments contains about 74k /
35k / 50k train / dev / test entities and 102 FIGER types.
Dev is used to optimize model hyperparameters. Appendix A gives some more statistics for MVET.

\textbf{Learning representation views}.
\seclabel{entityrep}
We refer to the three instances of the
representation metaview (see \secref{intro}) as 
\textbf{CTXT} (contexts),
\textbf{NAME} (name)
and \textbf{DESC} (description).

For learning CTXT embeddings, we train \textsc{wang2vec} \cite{ling15embeddings} on
Wikipedia after having replaced hyperlinked mentions of an
entity with its ID. 
NAME is derived from publicly available
300-dimensional fastText \cite{subword16} embeddings.  We
use the average of the words in a name as its NAME
embedding\footnote{Some of the Wikipedia titles contain a
  category inside parentheses, e.g.., ``Washington
  (state)''. We remove these parentheses and their content
  from the titles, if they exist, and then use the titles as our  names.}.  If a word does not have a fastText embedding,
we apply the fastText model to compute it.  So there are no
unknown words in our dataset.  For DESC, we extract the
keywords (using tf-idf) of the first paragraph of the Wikipedia article of
an entity. The DESC embedding is the average fastText embedding of the keywords.

To reiterate the complementarity of the three representation views:
names are ambiguous, but if we use the names of an entity in
different languages, we can mitigate this ambiguity.
E.g., ``Apple'' can refer to an entity or a fruit in
English, but only to an entity in French.
Similarly,
the description of an
entity is a high quality textual source to extract
information from.
The simplest case of complementarity  is that not
all views are available. An entity can be
completely missing from one of the languages; it may not
have a description because only a stub is provided; etc.

\subsection{Results} 
\seclabel{results}
\textbf{Evaluation metric.}  Following prior work in entity typing
\cite{figment17}, we evaluate by micro $F_1$,  a
global summary score of all system predictions.
Entity frequency is an important variable, so
we report results for tail (frequency $<$10, $n$=35,533),
head (frequency $>$100, $n$=2,638) and all entities.

\tabref{multiview} shows  results
for entity typing on our dataset, MVET.
We start with FIGMENT \cite{figment17} baseline results on MVET dataset (line 0), which
is the state-of-the-art system in entity typing. FIGMENT is equivalent to 
our MULTIVIEW-CON model with only English-CTXT, -NAME and -DESC 
 representations.

Lines 1--4, 9--12, 17--20 are singleview
results, e.g., $F_1$ for tail is 62.0 for the English-CTXT
view.
Lines 5--8, 13--16, 21--24 combine the four languages; so
these are multiview results for the language metaview. All
four multiview models are better than the corresponding singleview models
in the same block. Lines 25--28 show results for the
combination of the two metaviews; a total of twelve views
are combined (four languages times three
representations). The multi-multi-view
models on lines 25--28 outperform all other results.
Comparing line 25  and FIGMENT (line 0), adding representations from three more languages result in .5\%, .4\%, .9\% improvements for all, tail and head entities.
Line 26 by using ATT improves the results further especially for the tail entities.
These results confirm the effectiveness of our contributions: adding language as a metaview, and using ATT instead of CON to combine multiple views. 

Lines 29--32 show results for using  NAME representations in six additional
languages:
AR, FR, IT, PL, PT, RU. $F_1$ is up to more than one percent
better than on lines 13--16. This demonstrates the benefit
of using more languages -- although the effect is limited
since only the long tail of entities can improve. 

Lines
33--36 vs.\ lines 25--28 make the same comparison
(NAME4 vs.\ NAME10) for multi-multi-view. 
By ATT, we get a small
improvement for tail, but not for head (line 34 vs.\ line 26). 
Apparently, there is
noise added by considering more languages and this hurts the results for
head entities.

A general tendency is that ATT performs better compared to MAX, AVG and CON as the number
of views increases (lines 30 and 34) and so
the average number of views without information (i.e., missing views) for an
entity increases. In contrast to MAX, ATT can combine
different views. In contrast to CON and AVG, ATT can ignore some of
them based on low attention weights.

\begin{table}[t]
\centering{
\footnotesize
{
\begin{tabular}{l@{\hspace{0.025cm}}r@{\hspace{0.05cm}}|l||ccc}
&& & all  & tail  & head \\\hline 
& 0 & FIGMENT & 88.1 & 87.4 & 89.1 \\
\hline \hline 
\multirow{7}{*}{\scriptsize\rotatebox{90}{CTXT}} 
& 1& EN & 71.7 & 62.0 & 88.5\\
& 2 & DE & 31.7 & 14.6 & 76.6\\
& 3 & ES & 20.3 & 6.1 & 67.5 \\
& 4 & FA & 09.3 & 04.5 & 42.3 \\

& 5 & MULTIVIEW-CON &\textbf{73.7} & \textbf{64.6} & 89.6 \\
& 6 & MULTIVIEW-ATT & 73.6 & 64.5 & 89.2 \\
& 7 & MULTIVIEW-MAX & \textbf{73.7} & 64.2 & \textbf{89.8} \\
& 8 & MULTIVIEW-AVG & 73.0 & 63.5 & 89.3 \\

\hline 
\multirow{9}{*}{\scriptsize\rotatebox{90}{NAME4}} 

& 9 & EN & 73.4  & 73.1 & 76.1 \\
& 10 & DE & 34.1 & 23.8 & 68.5  \\
& 11 & ES & 29.8 & 20.4 & 65.3\\
& 12 & FA & 17.2 & 13.0 & 47.3 \\

& 13 & MULTIVIEW-CON & 75.8 & 75.0 & 81.0 \\
& 14 & MULTIVIEW-ATT & \textbf{76.1} & \textbf{75.3} & \textbf{81.2}\\
& 15 & MULTIVIEW-MAX  & 75.9 & 75.2 & 81.0 \\
& 16 & MULTIVIEW-AVG  & 75.8 & 75.1 & 80.9 \\

\hline 
\multirow{7}{*}{\scriptsize\rotatebox{90}{DESC}} 
& 17 & EN & 77.6 & 79.5 & 67.1\\
& 18 & DE & 28.9 & 20.6 & 51.4\\
& 19 & ES & 23.1 & 16.2 & 47.0 \\
& 20 & FA & 12.2 & 10.7 & 29.3 \\

& 21 & MULTIVIEW-CON & \textbf{81.6} & \textbf{82.3} & \textbf{78.2} \\
& 22 & MULTIVIEW-ATT & 79.6 & 80.6 & 73.8 \\
& 23 & MULTIVIEW-MAX & 79.5 & 80.5 & 74.4 \\
& 24 & MULTIVIEW-AVG & 78.9 & 79.7 & 75.2 \\

\hline 
\multirow{4}{*}{\scriptsize\rotatebox{90}{\begin{tabular}{c}CTXT\\ +NAME4\\ +DESC\end{tabular}}} 
& 25 & MULTIVIEW-CON & 88.6 & 87.8 & \textbf{90.0} \\
& 26 & MULTIVIEW-ATT & \textbf{89.0} & \textbf{88.3} & 89.8 \\
& 27 & MULTIVIEW-MAX & 87.9 & 86.9 & 89.6 \\
& 28 & MULTIVIEW-AVG & 87.1 & 86.2 & 88.4 \\

\hline 
\multirow{4}{*}{\scriptsize\rotatebox{90}{\begin{tabular}{c}NAME10\end{tabular}}} 
& 29 & MULTIVIEW-CON & 76.7 & 75.8 & 82.4 \\
& 30 & MULTIVIEW-ATT & \textbf{77.3} & \textbf{76.4} & \textbf{82.5} \\
& 31 & MULTIVIEW-MAX & 77.0 & 76.1 & 81.9 \\
& 32 & MULTIVIEW-AVG & 76.3 & 75.4 & 82.0 \\

\hline 
\multirow{4}{*}{\scriptsize\rotatebox{90}{\begin{tabular}{c}CTXT\\+NAME10\\+DESC\end{tabular}}} 
& 33 & MULTIVIEW-CON & 88.5 & 87.7 & \textbf{89.9} \\
& 34 & MULTIVIEW-ATT & \textbf{89.2} & \textbf{88.6} & 89.8 \\
& 35 & MULTIVIEW-MAX & 87.6 & 86.5 & 89.4 \\
& 36 & MULTIVIEW-AVG & 86.7 & 85.7 & 88.1 \\

\end{tabular} 
}
}
\caption{Micro $F_1$ for entity typing.
NAME4/NAME10 = name embeddings from 4 or 10 languages.\tablabel{multiview}
}
\end{table}

\begin{table}[t]
  \footnotesize
  
\begin{tabular}{l|@{\hspace{0.12cm}}c@{\hspace{0.05cm}}c@{\hspace{0.00cm}}|@{\hspace{0.12cm}}c@{\hspace{0.05cm}}c@{\hspace{0.00cm}}|@{\hspace{0.12cm}}c@{\hspace{0.05cm}}c}
& \multicolumn{2}{|c|}{CTXT} &
  \multicolumn{2}{|c|}{NAME} &
  \multicolumn{2}{|c}{DESC}\\
& SINGLE & CROSS 
& SINGLE & CROSS 
& SINGLE & CROSS \\ 

\hline 
EN & \textbf{71.7} & 71.6 &  \textbf{73.4} & 72.8 & 77.6 &  \textbf{82.0}\\
DE & 31.7 &\textbf{32.3} & 34.1 & \textbf{34.7} & 28.9 & \textbf{35.7}\\
ES & 20.3 & \textbf{21.2} & 29.8 & \textbf{30.7} & 23.1 & \textbf{28.6} \\
FA & 9.3 & \textbf{9.5} &  17.2 & \textbf{18.3} & 12.2 & \textbf{15.3}

\end{tabular} 

\caption{\tablabel{cross}Micro $F_1$ (all entities)  for
  twelve singleview models (SINGLE) and
one crossview model (CROSS)}
\end{table}

\subsection{Analysis: Crossview Learning}
To analyze whether sharing parameters across views is
important, \tabref{cross} compares 
(i) SINGLE: twelve different singleview models with
(ii) CROSS: a single crossview model that is trained on a training
set that combines the twelve individual singleview training
sets.
For CROSS, we use 
 \eqref{sharedspaceattention} with view specific transformation matrices,
mapping views in different spaces into a
common space, and then \eqref{mlp} with shared parameters across views.
The number of parameters in application is the same for
SINGLE and CROSS.

\tabref{cross} shows consistent and clear improvements of
CROSS  compared to SINGLE, except for
English CTXT and NAME. The English Wikipedia is
much larger than the others, so that embeddings based on it
have high quality. 
But our results
demonstrate that
training one model with
common parameters over all inputs is helping the
classification for non high-resource views.

Multiview learning exploits the complementarity of views: if
an entity's type cannot be inferred from one view, then
other views may have the required
information. \tabref{cross} shows
that using multiple views has a second
beneficial effect: even if applied to a single view, a model
trained on multiple views performs
better. \newcite{kan2016}'s image recognition and \newcite{Pappas_IJCNLP_2017}'s document classification findings
are similar.
Thus,
not only does the increased amount of available information
boost performance in the multiview setup, but also we can enable crossview transfer and learn a model that makes better predictions even if
information is only available from a single view.

\section{Related Work}
\textbf{Entity and mention typing}.
In this work, we assume that a predefined set of fine-grained types is given.
Entity typing, i.e.,  predicting types of a knowledge base entity
\cite{neelakantan2015inferring,figment15}, is the focus of this paper.
Mention typing, i.e., predicting types of a mention in a particular context
\cite{ling2012fine,RabinovichK17,shimaoka17eacl,murty2018hierarchical}, is a related task.
Mention typing models can be evaluated for entity typing when aggregating their predictions \cite{figment15,noise17,figment18jair}.
Therefore, our public and large entity typing dataset, MVET, can be used as an alternative to the small manually annotated mention typing datasets like the commonly used FIGER \cite{ling2012fine}.
We leave this to the future work.

\textbf{Multilingual entity typing.}
We build multilingual dataset and models for entity typing.
Most work on entity typing has been monolingual; e.g.,
 \newcite{figment15} (English); and
\newcite{Suzuki2016et} (Japanese).
There is work on mention typing \cite{erpv17multilingualFGEM}. 
\newcite{lin2017multiatt} that uses mono- and crosslingual
attention for relation extraction.
Crosslingual entity linking is an important related task,
where the task is to link mentions of entities in multilingual text to a knowledge base \cite{tsai2016cross}.
Many entities are not sufficiently annotated in Wikipedia, and therefore crosslingual entity linking is necessary to learn informative context representations from multiple languages.

\textbf{Multi-representation of entities}.
Aggregating information from multiple sources to learn
entity representations
has been explored
for entity typing \cite{figment17,figment18jair}, entity linking \cite{gupta2017emnlp}
and relation extraction \cite{wang16ijcai}.
Here, we
add language as a new ``dimension'' to
multi-representations: each language contributes a
different CTXT, NAME and DESC representation.

Our multilingual and multi-representation models are examples of
\textbf{multiview learning}.
\newcite{xu2013survey} and \newcite{zhao2017multi}
review the literature
on multiview learning.
\newcite{amini2009} cast
multilingual text classification, a task related to entity typing,
as  multiview learning.
\newcite{attMultiview2017}
address node classification and link prediction by
attention-based multiview representations
of graph nodes.
We also adopt a similar approach in our multiview representations for entity typing.

\section{Conclusion}
We formalized entity typing as a multiview problem by
introducing two metaviews, language and representation; each
combination of their instances defines a
distinct view.  Our experiments showed the effectivess
of this formalization by outperforming the state-of-the-art model.
Our basic idea of metaview learning is general and is applicable
to  related tasks, e.g., to relation extraction.
We release a large and public multiview and, in particular,
multilingual, entity typing dataset.

\appendix
\section{MVET Dataset Statistics}
\begin{table}[th]
\centering
\small
\begin{tabular}{r|r|r|r}
• & NAME & CTXT & DESC \\ 
\hline 
EN & 160,083 & 121,004 & 159,458 \\ 
\hline 
DE & 50,853 & 37,054 & 45,516 \\ 
\hline 
ES & 42,279 & 24,528 & 37,420 \\ 
\hline 
FA & 23,389 & 11,725 & 18,735 \\ 
\hline 
RU & 37,233 & 0 & 0 \\ 
\hline
FR & 54,434 & 0 & 0 \\ 
\hline
AR & 18,379 & 0 & 0 \\ 
\hline
PT & 31,879 & 0 & 0 \\ 
\hline
PL & 35,675 & 0 & 0 \\ 
\hline
IT & 41,686 & 0 & 0
\end{tabular} 
\caption{The statistics of our MVET dataset for each language and representation view.
The total number of entities is 160,083 with 102 types.}
\tablabel{stats}
\end{table}

\section*{Acknowledgments}
We thank Ehsaneddin Asgari, Katharina Kann, Valentina Pyatkin, T.J. Hazen and the anonymous reviewers for their helpful feedback on earlier drafts.
This work was partially supported by the European Research Council,
Advanced Grant NonSequeToR \# 740516.

\bibliography{ref}
\bibliographystyle{acl_natbib_nourl}

\newpage

\end{document}